\documentclass{article}

\usepackage{url}
\usepackage{subfig}
\usepackage{graphicx}

{}
{}
{}

\newcommand{\footremember}[2]{%
    \footnote{#2}
    \newcounter{#1}
    \setcounter{#1}{\value{footnote}}%
}
\newcommand{\footrecall}[1]{%
    \footnotemark[\value{#1}]%
}

\begin{document}

%\supertitle{Submission Template for IET Research Journal Papers}

\title{Employing Fusion of Learned and Handcrafted Features for Unconstrained Ear  Recognition}

\author{%
  Earnest E. Hansley\footremember{usf}{Computer Science and Engineering, University of South Florida, 4202 E. Fowler Avenue, ENB 118, Tampa, FL 33620, USA}%
  \and Maur\'icio Pamplona Segundo\footrecall{usf} \footnote{Department of Computer Science, Federal University of Bahia, Av. Adhemar de Barros S/N, IM 214, Salvador, BA 40170-110, Brazil}%
  \and Sudeep Sarkar\footrecall{usf} \footnote{sarkar@cse.usf.edu}%
  }

%\author{Earnest E. Hansley$^{1}$, Maur\'icio Pamplona Segundo$^{1,2}$, Sudeep Sarkar$^{1*}$}
%\affiliation{$^{1}$ Computer Science and Engineering, University of South Florida, 4202 E. Fowler Avenue, ENB 118, Tampa, FL 33620, USA\\
%$^{2}$ Department of Computer Science, Federal University of Bahia, Av. Adhemar de Barros S/N, IM 214, Salvador, BA 40170-110, Brazil\\
%$^{*}$ sarkar@cse.usf.edu}

%\author{\au{Earnest E. Hansley$^{1}$}, \au{Maur\'icio Pamplona Segundo$^{1,2}$}, \au{Sudeep Sarkar$^{1\corr}$}}
%\address{\add{1}{Computer Science and Engineering, University of South Florida, 4202 E. Fowler Avenue, ENB 118, Tampa, FL 33620, USA}
%\add{2}{Department of Computer Science, Federal University of Bahia, Av. Adhemar de Barros S/N, IM 214, Salvador, BA 40170-110, Brazil}
%\email{sarkar@cse.usf.edu}}

\maketitle

\begin{abstract}
We present an unconstrained ear recognition framework that outperforms state-of-the-art systems in different publicly available image databases. To this end, we developed CNN-based solutions for ear normalization and description, we used well-known handcrafted descriptors, and we fused learned and handcrafted features to improve recognition. We designed a two-stage landmark detector that successfully worked under untrained scenarios. We used the results generated to perform a geometric image normalization that boosted the performance of all evaluated descriptors. Our CNN descriptor outperformed other CNN-based works in the literature, specially in more difficult scenarios. The fusion of learned and handcrafted matchers appears to be complementary as it achieved the best performance in all experiments. The obtained results outperformed all other reported results for the UERC challenge, which contains the most difficult database nowadays. 
\end{abstract}

%%%%%%%%%%%%%%%%%%%%
%%% INTRODUCTION %%%
%%%%%%%%%%%%%%%%%%%%
\section{Introduction}

A number of researchers have shown that ear recognition is a viable alternative to more common biometrics such as fingerprint, face and iris~\cite{Chang2003,Kumar2012,Yan2005}. The ear is stable over time, less invasive to capture, and does not require as much control during image acquisition as other biometrics.  And, it is reasonable to assert that the ear has less privacy issues when compared to faces.  

\begin{figure}[!ht]
\centering
\includegraphics[height=1.7in]{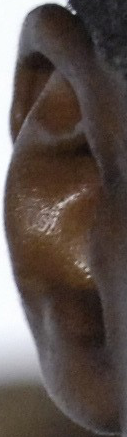}\hfill
\includegraphics[height=1.7in]{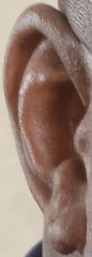}\hfill
\includegraphics[height=1.7in]{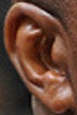}\hfill
\includegraphics[height=1.7in]{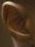}\hfill
\includegraphics[height=1.7in]{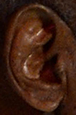}
\caption{Example of a challenging task for ear recognition in an unconstrained setting: given five images of four different subjects, can you tell which pair of images belong to the same person?}
\label{fig1}
\end{figure}

Traditionally ear recognition research has been performed on ear images that were captured in an ideal setting.  In an ideal setting, the ears are all captured in the same position, with the identical lighting, and identical resolution. With the advances in computer vision and pattern recognition techniques, research of ear recognition is shifting to a more challenging scenario whereby ear images are acquired from real world (unconstrained) settings~\cite{Emersic2017,Emersic2017b}. It is more difficult to recognize ears in the wild.  In this paper we use "ears in the wild" and "unconstrained ears" interchangeably.  Figure~\ref{fig1} illustrates the difficulty of recognizing individuals through ear images in the wild.

This example mostly illustrates the problem of pose variation, but many other factors may affect the recognition performance: different acquisition devices, low resolution, illumination variations, occlusions caused by hair and head accessories, earrings, ear plugs and so on. To overcome these recognition challenges, ear recognition has to achieve good results for non-cooperative subjects.  This will make ear biometric recognition very useful for practical purposes, like video surveillance and continuous authentication.

In order to carry out the task of recognizing humans through their ears, a common sequence of steps is usually followed (see Figure~\ref{fig_diagram}):

\begin{figure}[!ht]
\centering
\includegraphics[width=12.0cm]{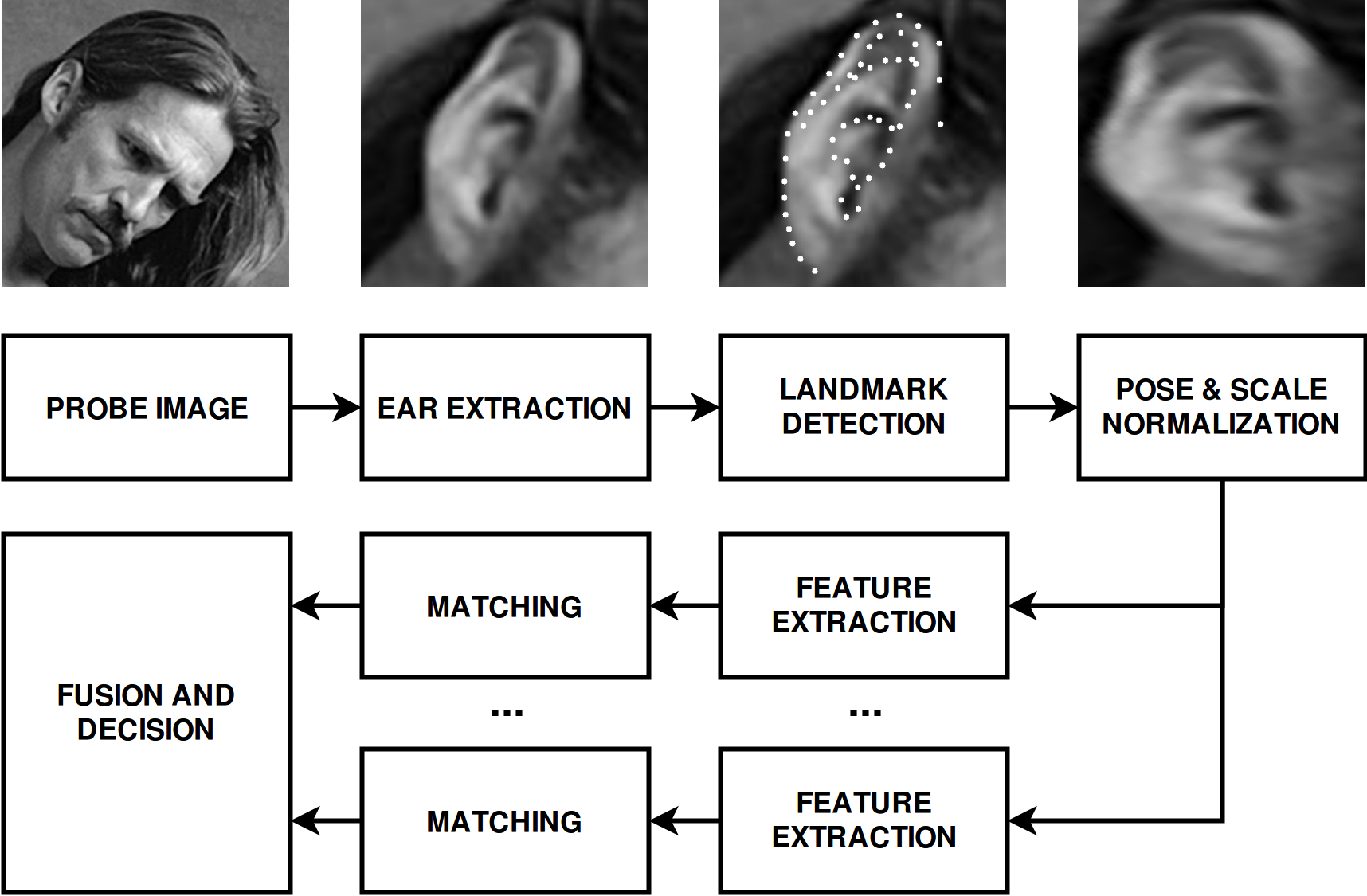}
\caption{Diagram of our ear recognition framework. Given an unconstrained image, ears are cropped using ground truth annotations.  Landmark detection is performed to obtain the information required to normalize pose and scale variations. Normalized images are described by different feature extractors and matched through distance metrics.  Scores are fused and a recognition decision is made.}
\label{fig_diagram}
\end{figure}

\begin{description}
\item[\bf Acquisition step:]~captures a digital biometric sample using an appropriate sensor. For all of our experiments we use images from five publicly available databases (Section~\ref{sec_databases}).
\item[\bf Localization step:]~locates the biometric information and separates it from existing irrelevant parts of the acquired sample. The images we used were either already cropped or the ground truth location of the ears was provided; thus we do not perform the localization step. However, it is possible to find successful approaches that perform ear detection in the wild in the literature~\cite{Zhang2017,Emersic2017c}.
\item[\bf Normalization step:]~reshapes the input sample to a standard format to reduce unwanted variations. We use a landmark detector based on Convolutional Neural Networks (CNN)~\cite{Krizhevsky2012} to locate a set of 55 landmarks (Section~\ref{sec_landmark}), which are then employed to translate, rotate and scale the input image to a standard configuration (Section~\ref{sec_normalization}).
\item[\bf Features Description step:]~selects discriminant features from a  normalized sample and usually reduces its dimensionality. We use a state-of-the-art CNN architecture for face recognition for the task of ear recognition in the wild (Section~\ref{sec_learned}), as well as different traditional ear description approaches (Sections~\ref{sec_pca}~and~\ref{sec_handcrafted}).
\item[\bf Recognition step:]~compares descriptors and decides whether they belong to the same person or not. All images are compared to each other using the descriptor's distance metric.  All scores are normalized using Min-max normalization~\cite{Jain2005}, then score level fusion~\cite{Kittler1998} is used to combine results of different descriptors and inform the decision. (Section~\ref{sec_fusion}).
\end{description}

After implementing our framework and evaluating the results, the following contributions were reached:

\begin{itemize}
\item we designed and developed a two-stage CNN-based landmark detector that achieves accurate results even in the presence of variations not seen in the training data (Section~\ref{sec_landmark}).  We use the detector to automatically normalize images and instantly observed a boost in the recognition rate; \\
\item we devised a CNN-based ear descriptor based on a state-of-the-art face recognition architecture that outperformed other state-of-the-art ear recognition works that are based on CNNs; \\
\item we showed that handcrafted and learned descriptors are complementary, and thus a considerable increase in performance can be reached when both are fused.
\end{itemize}

%%%%%%%%%%%%%%%%%
%%% DATABASES %%%
%%%%%%%%%%%%%%%%%
\section{Databases}
\label{sec_databases}

There are many things that can affect the performance of ear recognition and some sets of ear images are easier than others.  Therefore, it is a good idea for researchers to test on multiple image datasets when feasible.  A test may include using ideal images then progressing to unconstrained more difficult images to recognize.  In this work we use five different databases to train and evaluate our ear recognition framework.  We use images from the Indian Institute of Technology Delhi Ear Database (IIT), the West Pomeranian University of Technology Ear Database (WPUTE), the Annotated Web Ears database (AWE), the In-the-wild Ear Database (ITWE) and the Unconstrained Ear Recognition Challenge database (UERC). More details about each of them are given in the subsequent sections.

\subsection{Indian Institute of Technology Delhi Ear Database}

\begin{figure}[!ht]
\centering
\subfloat[]{\label{fig_iit}\includegraphics[height=0.84in]{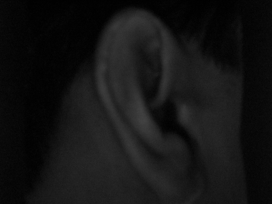}\includegraphics[height=0.84in]{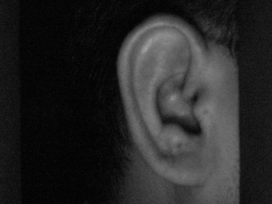}\includegraphics[height=0.84in]{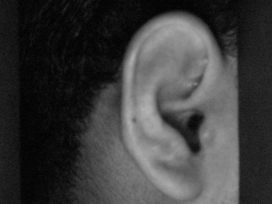}} \\
\subfloat[]{\label{fig_wpute}\includegraphics[height=1.1in]{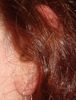}\includegraphics[height=1.1in]{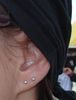}\includegraphics[height=1.1in]{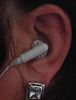}\includegraphics[height=1.1in]{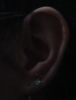}} \\
\subfloat[]{\label{fig_awe}\includegraphics[height=1.44in]{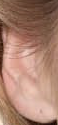}\includegraphics[height=1.44in]{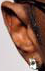}\includegraphics[height=1.44in]{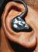}\includegraphics[height=1.44in]{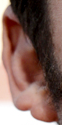}} \\
\subfloat[]{\label{fig_itwe}\includegraphics[height=1.2in]{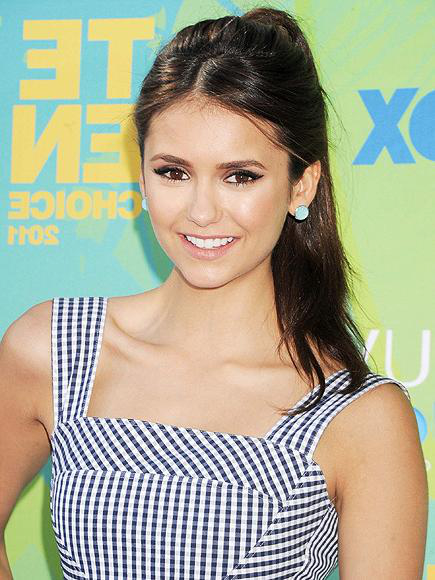}\includegraphics[height=1.2in]{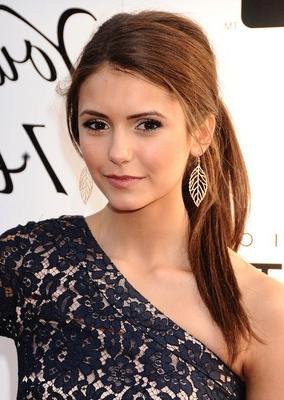}\includegraphics[height=1.2in]{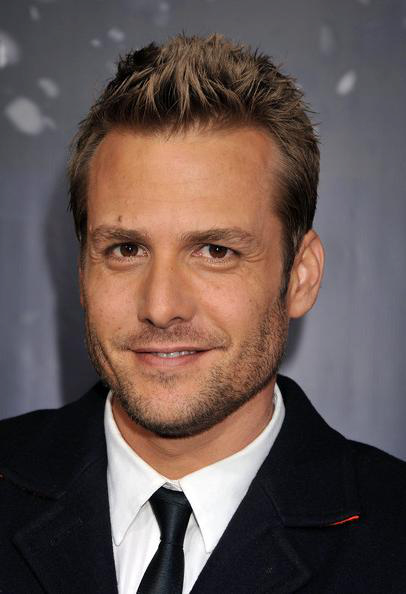}\includegraphics[height=1.2in]{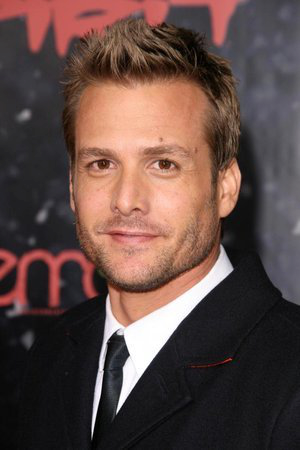}}
\caption{Example of images from (a) IIT, (b) WPUTE, (c) AWE and (d) ITWE databases. While all IIT images have a similar controlled disposition ({\it i.e.} pose, resolution, illumination), the other three databases are unconstrained and present different challenges for recognition: occlusions caused by hair and head accessories, ear plugs, earrings, variations in illumination and pose. The AWE and ITWE images have the most variation in terms of resolution and age. The WPUTE images on the other hand have same subject images that were mostly acquired in a single session and have the same size.}
\label{fig_databases}
\end{figure}

The IIT database~\cite{Kumar2012} was released in two different formats, a raw version and a normalized version. We use the raw version for our experiments.  It contains 493 images with size $272\times204$ from 125 different subjects. Each image shows a small region around the left ear and was collected in an indoor environment in a well-controlled acquisition setup, which makes this database a suitable benchmark for a nearly ideal ear recognition scenario. Figure~\ref{fig_iit} shows some raw images provided by the IIT database.

\subsection{West Pomeranian University of Technology Ear Database}

The WPUTE database~\cite{Frejlichowski2010} was originally created to evaluate the performance of ear recognition in the wild. It contains images that state-of-the-art ear recognition approaches could not handle at that time. The images reflect the challenges associated with ear recognition, such as occlusions caused by hair, earrings, and ear plugs. The database also provides images with variations in gender, ethnicity, pose, illumination and acquisition sensor. However, because the vast majority of images from a same person were acquired in a single session, intraclass variation is minimal. Thus, although the preprocessing step was heavily affected by these variations, some of the variations could in fact benefit the recognition task ({\it e.g.} a person wearing the same earring in all acquisitions). This database provides 3348 images with size $380\times500$ from 474 different subjects ({\it i.e.} each subject has at least 4 images) showing a small region around the ear. However, 1388 of them are duplicates, which may have inflated the reported accuracy of some works in the literature~\cite{Zhou2017}, and we also found 6 images that were mistakenly labeled as left ears while they actually were right ears. After removing duplicates and fixing labels, 1960 images are available for use, 982 from left ears and 978 from right ears. Some examples of WPUTE images are shown in Figure~\ref{fig_wpute}.

\subsection{Annotated Web Ears database}

The AWE database~\cite{Emersic2017} contains 1000 images from 100 different subjects ({\it i.e.} 10 images per subject) which were collected from searches for public figures on the Internet. Image size varies from $15\times29$ to $473\times1022$ pixels, with size $83\times160$ on average. Ears were tightly cropped, so the proportion of background pixels is the smallest among all databases used in this work. All variations presented in the WPUTEDB database are also present in the AWE database in an intenser form. Although it labels ears as left and right, with 520 and 480 images respectively, the images may have been inadvertently flipped horizontally before being released on the Internet. So, it is possible that there are some noisy labels. Some of the challenges encountered in the AWE database are exemplified in Figure~\ref{fig_awe}.

\subsection{In-the-wild Ear Database}

The ITWE database~\cite{Zhou2017} is divided in two sets, Collection A and Collection B. Collection A was collected using Google image search and contains 605 images without identity reference, but with 55 manually annotated landmarks. The position of these landmarks can be observed in Figure~\ref{fig_diagram}. This collection was randomly split in a training set with 500 images and a test set with 105 images. It is suitable for training ear detection and normalization approaches, but not for recognition purposes. For this reason, Collection B was created for recognition evaluation and contains 2058 images from 231 different subjects taken from three public databases for face recognition in the wild: VGG-Face~\cite{Parkhi2015}, LFW~\cite{Huang2007} and Helen Dataset~\cite{Le2012}. Bounding boxes for each ear were obtained by a detector based on histograms of oriented gradients (HOG)~\cite{Dalal2005} which was trained on images from Collection A, and these box coordinates were released together with this collection. Images in both Collection A and Collection B include cluttered backgrounds ({\it e.g.} face, body parts, scenario) and vary considerably in size and ear resolution. Variations in ear images of the ITWE database are comparable to the AWE database ones, but there is no differentiation between left and right ears ({\it i.e.}.  The ITWE images are horizontally flipped so that all have the same orientation), which is a problem for recognizing people with asymmetric ears ({\it i.e.} about 10\% of people according to Yan~and~Bowyer~\cite{Yan2005}). In addition, we were able to find many mislabeled samples.  We did not fix any of them for comparison purposes. Some examples of ITWE images are presented in Figure~\ref{fig_itwe}.

\subsection{Unconstrained Ear Recognition Challenge database}

The UERC database~\cite{Emersic2017b} is an extension of the AWE database and was built for competition purposes. The major change is the amount of images and subjects, which increased considerably. The database is divided in two parts, with 2304 images from 166 subjects for training and 9500 images from 3540 subjects for testing. The subjects designated for training have at least 10 images, while subjects in test may contain only one image. A portion of the subjects in training and testing ({\it i.e.} 150 and 180 subjects, respectively) have exactly 10 images. Ears may be left or right oriented, but ground truth annotations of the orientation are only available for training images.

\subsection{Discussion}

The five sets of images we use have different levels of difficulty so that we can conduct a fair test and evaluate the performance of our framework implementation, then compare our results to the state-of-the-art. While the IIT ear images are not unconstrained, they can be used to detect overfitting to the wild scenarios ({\it i.e.} using images from easier databases should always result in higher accuracy), a problem that was already observed in works that recognize faces in the wild~\cite{Dahia2017}. Although all the remaining databases are unconstrained, based on their descriptive characteristics, we conclude that WPUTE and UERC are respectively the least and the most challenging unconstrained image sets, while AWE and ITWE have a similar difficulty level.

%%%%%%%%%%%%%%%%%%%%%%%%%%
%%% LANDMARK DETECTION %%%
%%%%%%%%%%%%%%%%%%%%%%%%%%
\section{Landmark detection}
\label{sec_landmark}

Even with the recent emergence of deep learning methods for biometric recognition in uncontrolled scenarios, normalization is still necessary to achieve better results. For instance, a landmark-based orientation and scale normalization is a standard procedure in face recognition state-of-the-art works~\cite{Wen2016,Wu2015}. With this in mind, we pursued a similar path for the ear recognition problem by investigating the use of CNNs for the landmark detection task. To this end, we use images and annotations provided in Collection A of the ITWE database for CNN training and accuracy evaluation. As only 500 images are available for training, we performed different data augmentation operations in order to avoid overfitting and increase the network generalization power. For each training image, we use principal component analysis (PCA)~\cite{Abdi2010} on the 2D coordinates of the annotated landmarks to obtain the upright orientation of the ear ({\it i.e.} we assume it corresponds to the direction of the first component). Then, we create multiple images by rotating the upright ear from $-45^\circ$ to $45^\circ$ with steps of $3^\circ$. Each ear is also transformed by a random scale change of up to 20\% of the original ear size in both axes, as well as a random translation of up to 20\% of the original ear size in each axis. After applying all these modifications, images were rescaled to 96$\times$96 pixels and we ended up with 15500 training images.

The architecture of our network is based on a common design nowadays, even for landmark detection~\cite{Sun2013}, which consists of alternating between convolution and max pooling layers in the beginning, and then following with a sequence of fully-connected layers. We use rectified linear units in convolution and fully-connected layers to train models from scratch.  We also added dropouts after all max pooling and the first fully-connected layers to avoid overfitting the training data. A complete description of our architecture is presented in Table~\ref{cnn-landmarks}. It was implemented using TensorFlow, and the optimization to minimize the mean squared error in the output was carried out by the Nesterov's Momentum algorithm~\cite{Sutskever2013} for 2000 epochs.

\begin{table}[!h]
\caption{Network architecture for landmark detection in ear images. It receives as input a grayscale image with 96$\times$96 pixels and outputs a 110-dimensional vector representing 2D coordinates for 55 predefined landmarks.\label{cnn-landmarks}}{
\centering
\small
\begin{tabular}{c|c|c|c|c|c|c}
\textbf{\#} & \textbf{Type} & \textbf{Input} & \textbf{Filter} & \textbf{Stride} & \textbf{Drop} & \textbf{Output} \\ \hline
1 & Conv/Relu  & 96$\times$96$\times$1   & 3$\times$3$\times$1$\times$32   & 1 &      & 96$\times$96$\times$32  \\
2 & MaxPool    & 96$\times$96$\times$32  & 2$\times$2                      & 2 & 10\% & 48$\times$48$\times$32  \\
3 & Conv/Relu  & 48$\times$48$\times$32  & 2$\times$2$\times$32$\times$64  & 1 &      & 48$\times$48$\times$64  \\
4 & MaxPool    & 48$\times$48$\times$64  & 2$\times$2                      & 2 & 20\% & 24$\times$24$\times$64  \\
5 & Conv/Relu  & 24$\times$24$\times$64  & 2$\times$2$\times$64$\times$128 & 1 &      & 24$\times$24$\times$128 \\
6 & MaxPool    & 24$\times$24$\times$128 & 2$\times$2                      & 2 & 30\% & 12$\times$12$\times$128 \\
  & Flattening & 12$\times$12$\times$128 &                                 &   &      & 18432                   \\
7 & Fc/Relu    & 18432                   &                                 &   & 50\% & 1000                    \\
8 & Fc/Relu    & 1000                    &                                 &   &      & 1000                    \\
9 & Fc         & 1000                    &                                 &   &      & 110            
\end{tabular}
}{}
\end{table}

Although this network achieved a good accuracy considering the level of variations in unconstrained scenarios, we evaluated a two-stage solution, whereby the first network is used to create an easier landmark detection scenario by reducing scale and translation variations, and the second network is used to generate the 2D coordinates for landmarks. To this end, we use the coordinates obtained by the network described above to refine the center and the orientation of an ear using PCA, and then feed the rectified image to a second network that was trained in a more controlled scenario. The second network has the same architecture and optimization procedure of the first one, the only difference is the training data, which uses less variation in the augmentation process.  Rotations are performed from $-15^\circ$ to $15^\circ$ with steps of $1^\circ$, and random scale and translation changes are limited to up to 10\% of the original ear size.

%%%%%%%%%%%%%%%%%%%%%
%%% NORMALIZATION %%%
%%%%%%%%%%%%%%%%%%%%%
\section{Geometric normalization}
\label{sec_normalization}

\begin{figure}[!ht]
\centering
\subfloat[]{\label{fig_norm1}\includegraphics[width=1.15in]{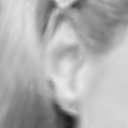}}\hfill
\subfloat[]{\label{fig_norm2}\includegraphics[width=1.15in]{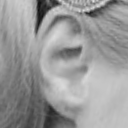}}\hfill
\subfloat[]{\label{fig_norm3}\includegraphics[width=1.15in]{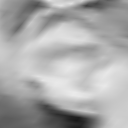}}\hfill
\subfloat[]{\label{fig_norm4}\includegraphics[width=1.15in]{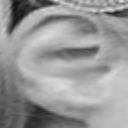}}
\caption{Normalization results (a)-(b) with and (c)-(d) without the same sampling rate in both axis for two ear images of the same person with pose variations. Different sampling rates are employed to reduce the difference in the width of the ear caused by pose variations. While (b) is 25\% wider than (a), the difference in width between (c) and (d) is negligible.}
\label{fig_norm}
\end{figure}

After landmark detection, we normalize the ears by applying PCA on the retrieved landmarks. We use the first component as the orientation of the ear and the center of the oriented bounding box as the center of the ear. We then interpolate a $128\times128$ image with these parameters considering that the distance between the center of the ear and the top of the image is equal to two times the squared root of the first eigenvalue in the original image. However, as ears in the wild may present significant pose variations, this also occurs in width variations that may affect the recognition performance, as shown in Figures~\ref{fig_norm1}~and~\ref{fig_norm2}. Thus, we use different sampling rates in x and y directions in a way that the distance between the center of the ear and one side of the image is equal to two times the squared root of the second eigenvalue in the original image. This way, the width and the height of the normalized ear are approximately the same, as may be seen in Figures~\ref{fig_norm3}~and~\ref{fig_norm4}, and image variations caused by pose become less intense.

%%%%%%%%%%%%%%%%%%%%%%%%%%%%%%%%%%%%
%%% EAR DESCRIPTION AND MATCHING %%%
%%%%%%%%%%%%%%%%%%%%%%%%%%%%%%%%%%%%
\section{Description and matching}
\label{sec_recognition}

We evaluate three different description and matching schemes based on 1) holistic image features, 2) handcrafted features and 3) learned features. We then investigate if fusing some of them can achieve a higher accuracy. More details are given in the following sections.

\subsection{Holistic features}
\label{sec_pca}

PCA was one of the first methods employed to the ear recognition problem~\cite{Chang2003}, as it provides a holistic description of the sample images while reducing the dimensionality of the data. However, even the pioneer works using PCA already reported a performance drop caused by variations in pose and illumination, and such variations are much more intense in recent uncontrolled databases. We used a PCA implementation available in the Face Identification Evaluation System~\cite{Yambor2002} as a baseline approach, and its feature vectors were matched though the Mahalanobis distance. The first 20 eigenvectors were dropped to avoid illumination and background variations, and we kept 60\% of the eigenvectors in our PCA descriptor.

\subsection{Handcrafted features}
\label{sec_handcrafted}

As holistic features are strongly affected by different variations, specialists designed different feature extraction approaches, which are known as handcrafted features, seeking to overcome some of these problems. Emersic~\emph{et~al.}~\cite{Emersic2017,Emersic2017b} released a toolbox that allows the extraction of the best performing state-of-the-art handcrafted features for ear recognition: local binary patterns (LBP), binarized statistical image features (BSIF), local phase quantization features (LPQ), rotation invariant LPQs (RILPQ), patterns of oriented edge magnitudes (POEM), HOG, dense scale-invariant feature transform (DSIFT) and Gabor wavelets. All descriptors were extracted using the default parameters of the toolbox. For matching, as in Emersic~\emph{et~al.}'s work~\cite{Emersic2017}, we compared histogram-based descriptors using the chi-square distance and Gabor descriptors using the cosine distance.

\subsection{Learned features}
\label{sec_learned}

Considering that the performance of handcrafted descriptors degrades when using uncontrolled ear images~\cite{Emersic2017}, we employed CNNs in so that we could boost improve performance and so that we could learn more about the images, as well as how to describe them in a more discriminative and concise way. The CNN that we implemented is a state-of-the-art CNN architecture employed for face recognition in the wild~\cite{Wen2016} and we trained it from scratch for the ear recognition in the wild problem. We present a complete description of the chosen CNN architecture as well as specific layer configurations in Table~\ref{cnn-recognition}. This network was also implemented using TensorFlow, and the Adam optimization algorithm~\cite{Kingma2014} was used to minimize the weighted sum of softmax and center losses. As in Wen~\emph{et~al.}'s~work~\cite{Wen2016}, we set the center loss weight to $0.003$.

\begin{table}[!h]
\caption{Network architecture for feature extraction in ear images. It receives as input a grayscale image with 128$\times$128 pixels and outputs a 512-dimensional vector containing a discriminative feature representation of the input image.\label{cnn-recognition}}{
\footnotesize
\begin{tabular}{c|c|c|c|c|c|c}
\textbf{\#} & \textbf{Type} & \textbf{Input} & \textbf{Filter} & \textbf{Stride} & \textbf{Drop} & \textbf{Output} \\ \hline
1  & Conv/Relu    & 128$\times$128$\times$1   & 3$\times$3$\times$1$\times$128   & 1 &      & 128$\times$128$\times$128 \\
2  & Conv/Relu    & 128$\times$128$\times$128 & 3$\times$3$\times$128$\times$128 & 1 &      & 128$\times$128$\times$128 \\
3  & MaxPool      & 128$\times$128$\times$128 & 2$\times$2                       & 2 & 10\% & 64$\times$64$\times$128   \\
4  & Conv/Relu    & 64$\times$64$\times$128   & 3$\times$3$\times$128$\times$128 & 1 &      & 64$\times$64$\times$128   \\
5  & MaxPool      & 64$\times$64$\times$128   & 2$\times$2                       & 2 & 20\% & 32$\times$32$\times$128   \\
6  & Conv/Relu    & 32$\times$32$\times$128   & 3$\times$3$\times$128$\times$256 & 1 &      & 32$\times$32$\times$256   \\
7  & MaxPool      & 32$\times$32$\times$256   & 2$\times$2                       & 2 & 30\% & 16$\times$16$\times$256   \\
8  & Conv/Relu    & 16$\times$16$\times$256   & 3$\times$3$\times$256$\times$256 & 1 &      & 16$\times$16$\times$256   \\
9  & MaxPool      & 16$\times$16$\times$256   & 2$\times$2                       & 2 &      & 8$\times$8$\times$256     \\
10 & Conv/Relu    & 8$\times$8$\times$256     & 3$\times$3$\times$256$\times$256 & 1 &      & 8$\times$8$\times$256     \\
   & Flattening 8 & 8$\times$8$\times$256     &                                  &   &      & 16384                     \\
   & Flattening 9 & 8$\times$8$\times$256     &                                  &   &      & 16384                     \\
   & Concat 8\&9  & 16384/16384               &                                  &   &      & 32768                     \\
11 & Fc           & 32768                     &                                  &   &      & 512                       \\        
\end{tabular}
}{}
\end{table}

\begin{figure}[!ht]
\centering

\includegraphics[width=3.9cm]{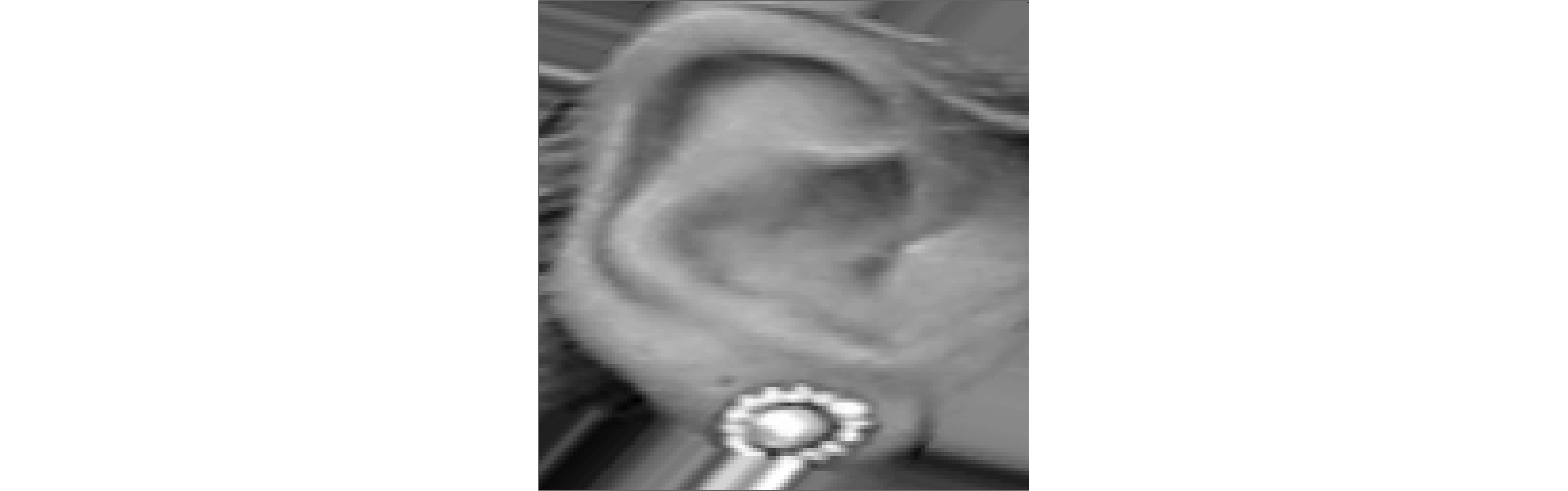}\hfill\includegraphics[width=3.9cm]{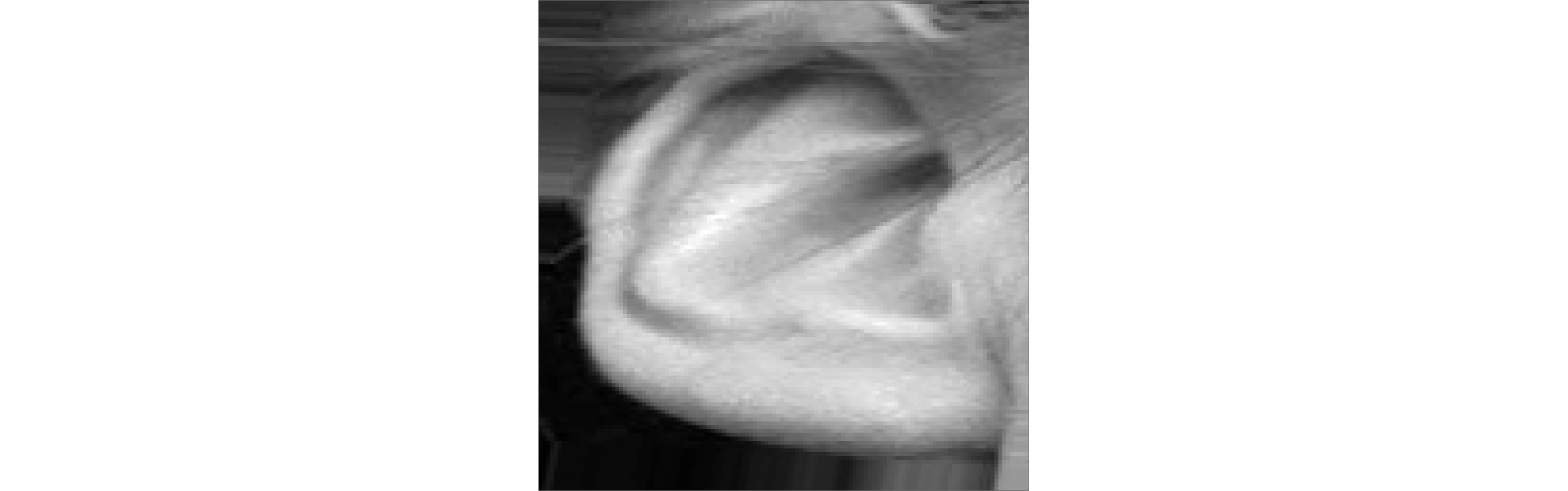}\hfill\includegraphics[width=3.9cm]{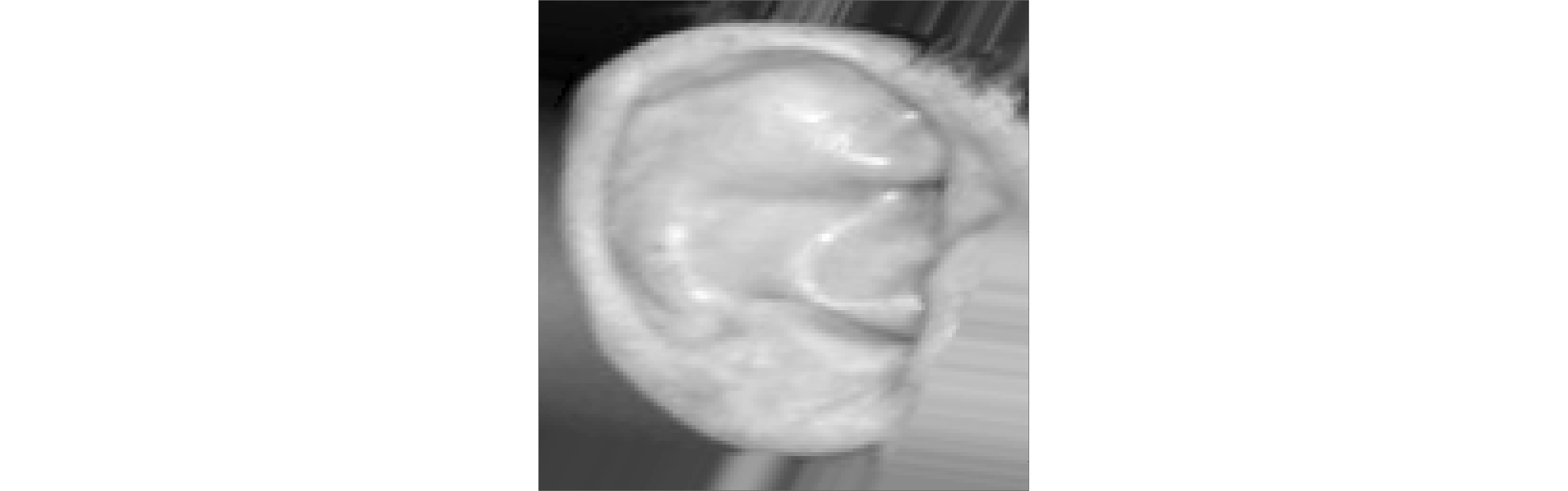}\\ \vspace{0.5cm}
\hfill1st MaxPool\hfill~~~~~\hfill1st MaxPool\hfill~~~~~\hfill1st MaxPool\hfill~\\ \vspace{0.3cm}
\includegraphics[width=3.9cm]{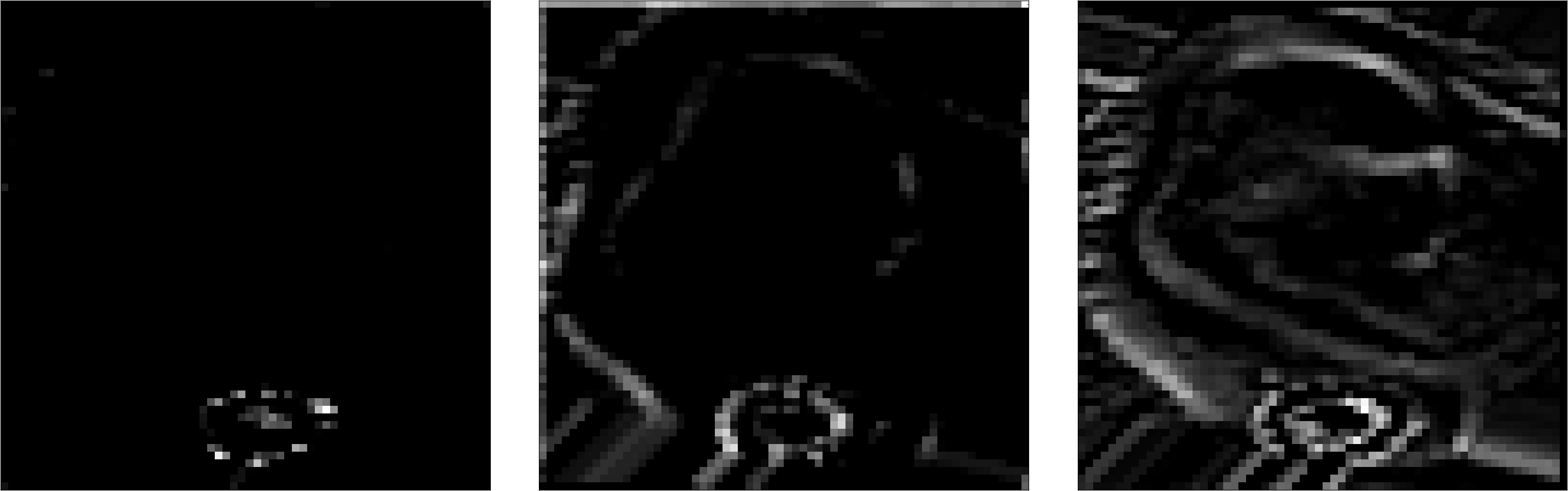}\hfill\includegraphics[width=3.9cm]{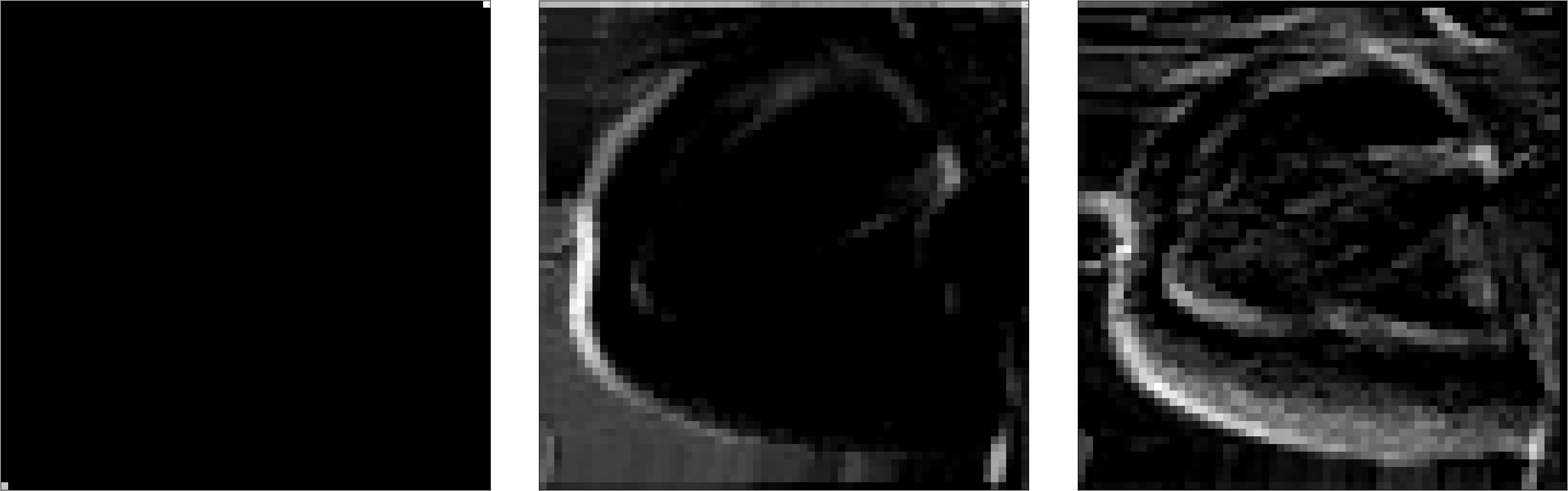}\hfill\includegraphics[width=3.9cm]{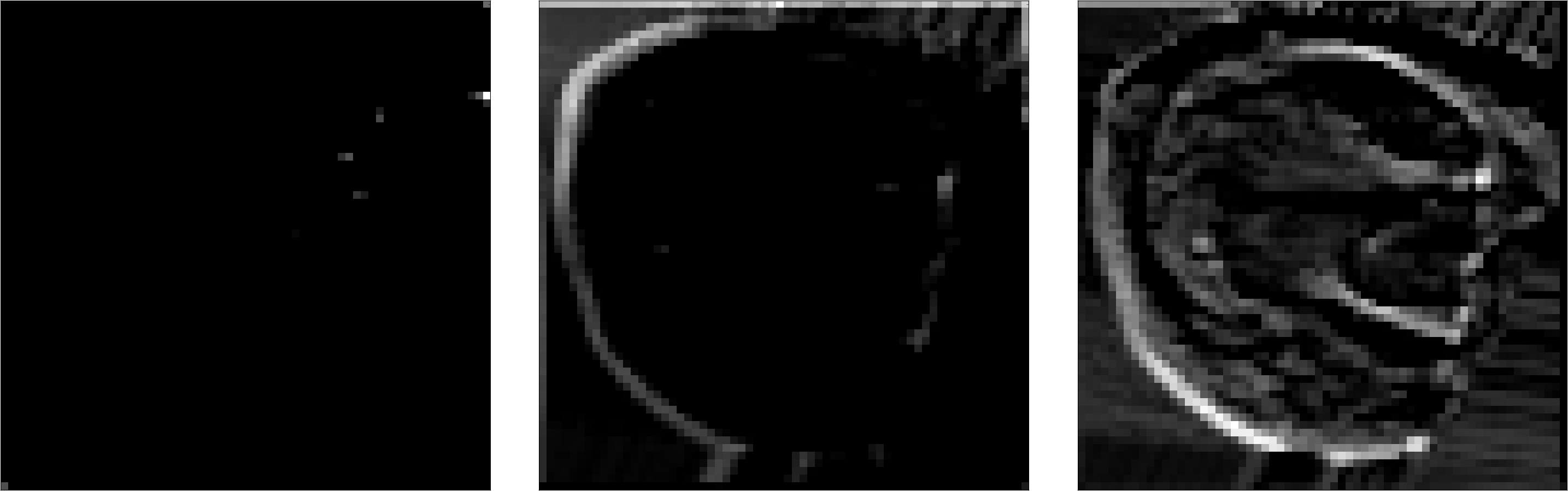}\\ \vspace{0.5cm}
\hfill2nd MaxPool\hfill~~~~~\hfill2nd MaxPool\hfill~~~~~\hfill2nd MaxPool\hfill~\\ \vspace{0.3cm}
\includegraphics[width=3.9cm]{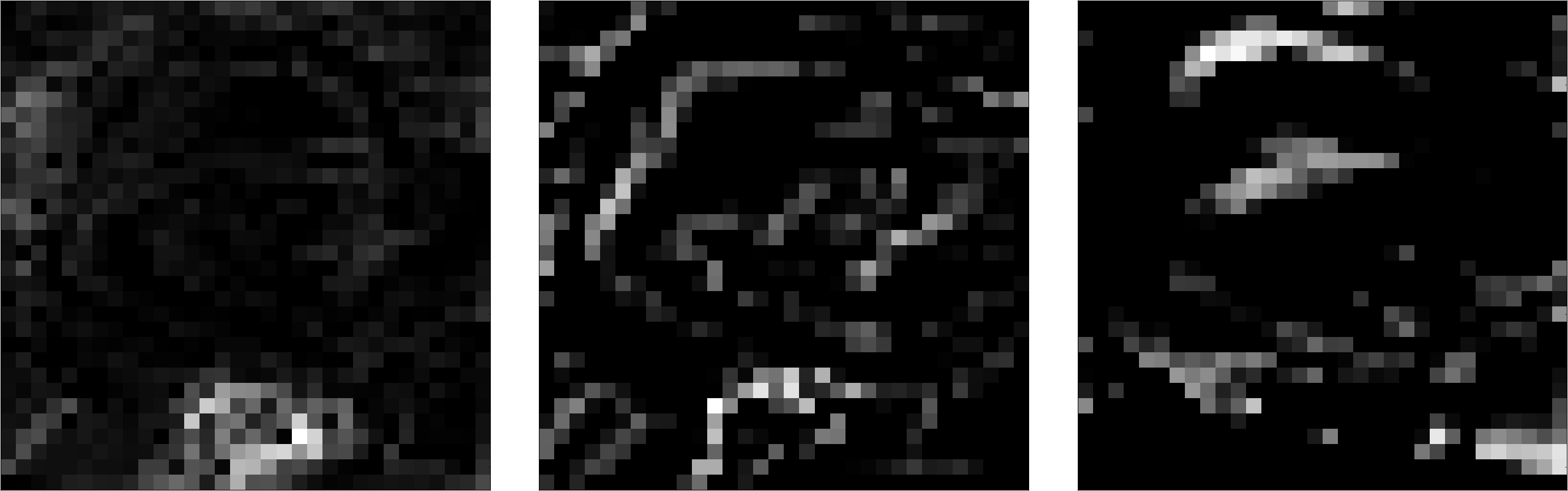}\hfill\includegraphics[width=3.9cm]{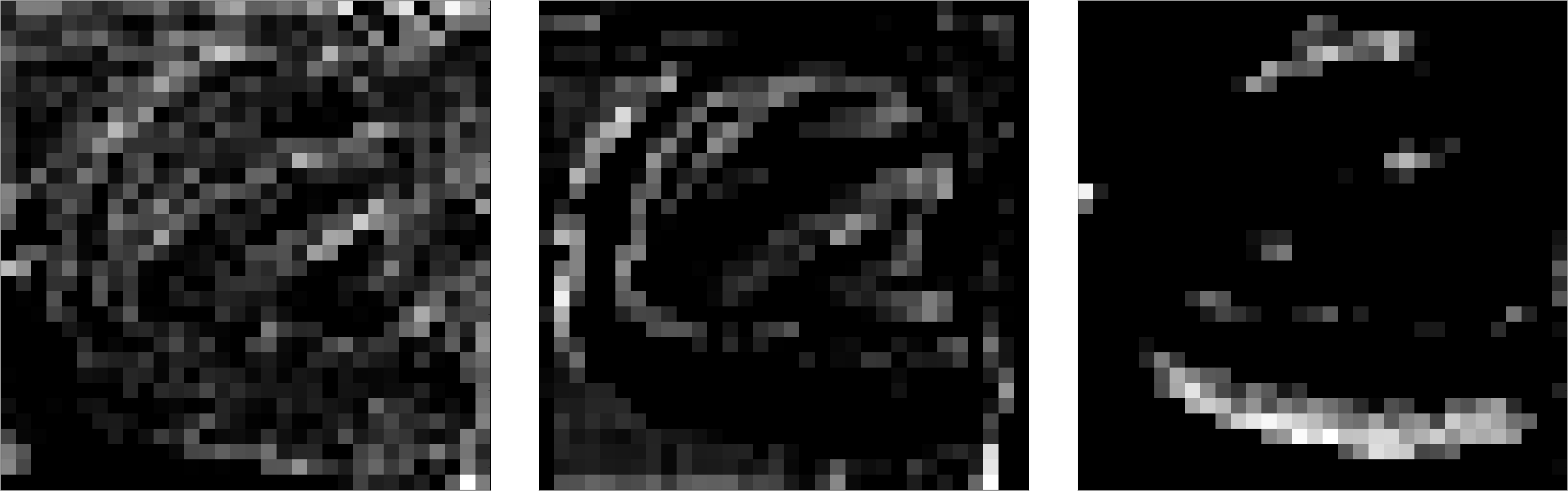}\hfill\includegraphics[width=3.9cm]{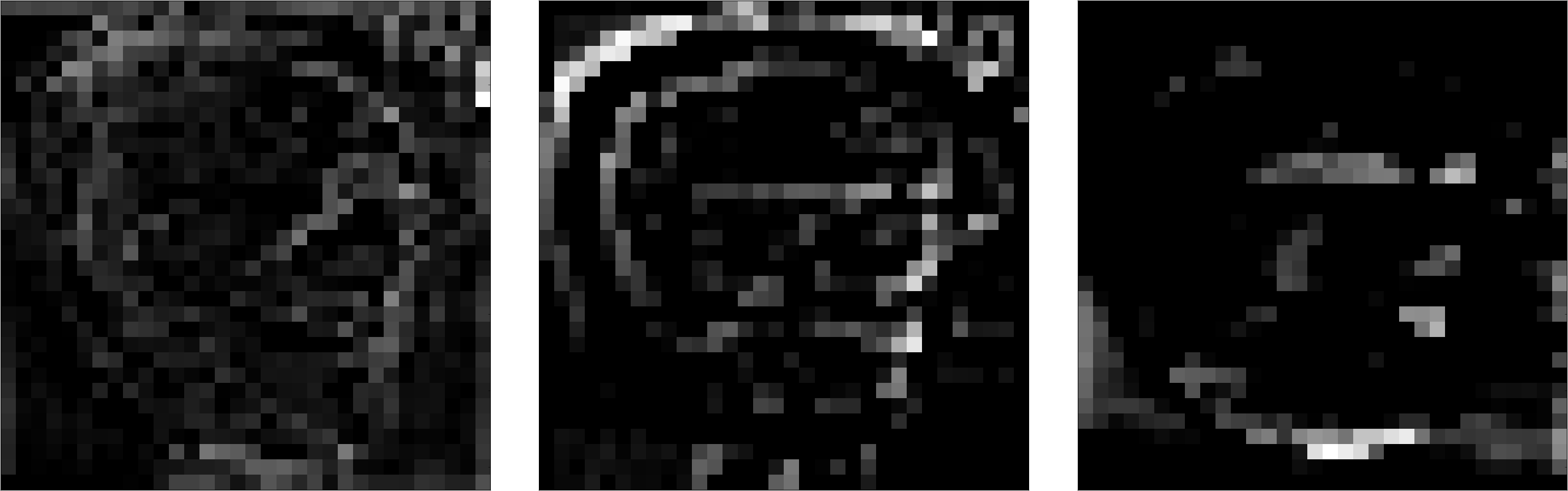}\\ \vspace{0.5cm}
\hfill3rd MaxPool\hfill~~~~~\hfill3rd MaxPool\hfill~~~~~\hfill3rd MaxPool\hfill~\\ \vspace{0.3cm}
\includegraphics[width=3.9cm]{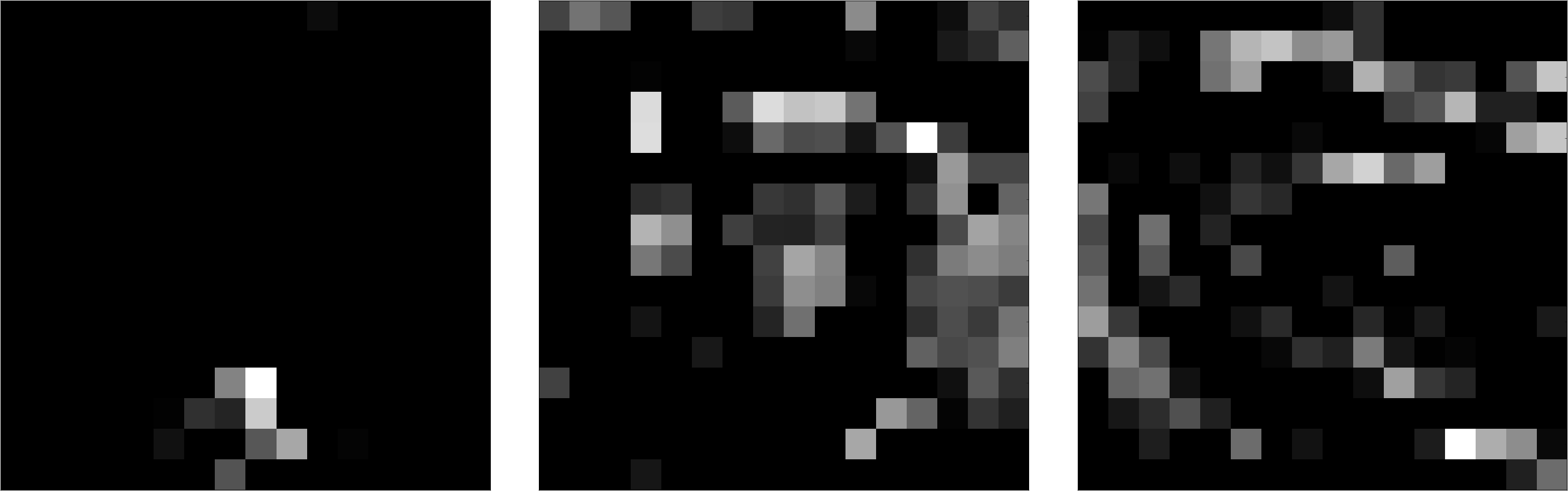}\hfill\includegraphics[width=3.9cm]{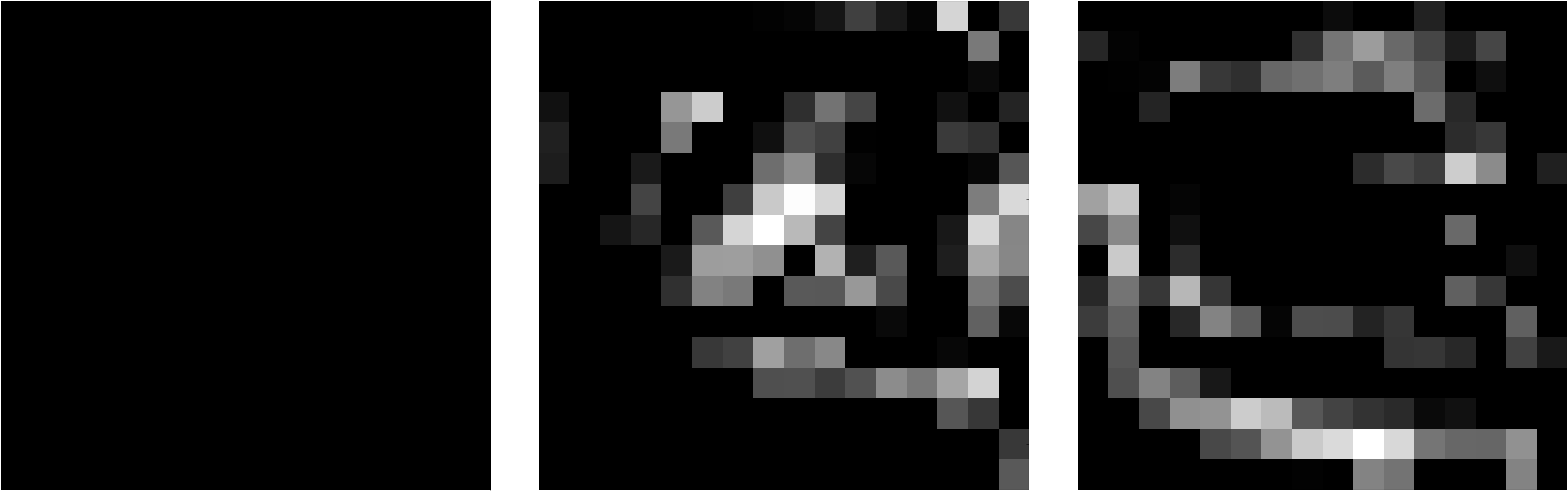}\hfill\includegraphics[width=3.9cm]{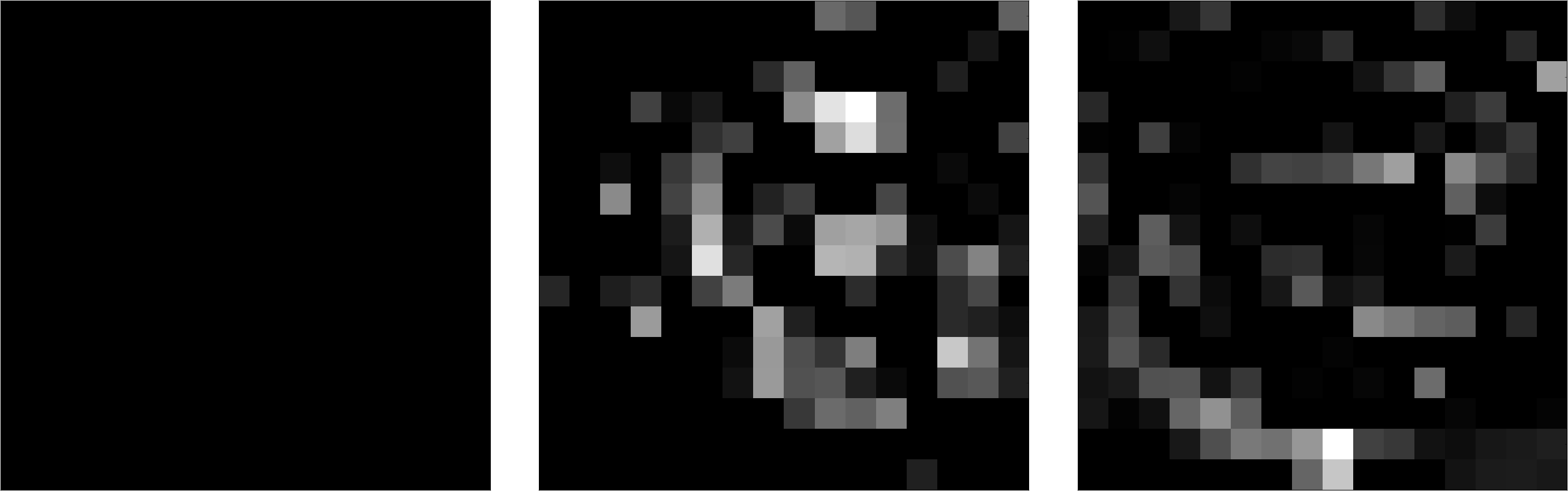}\\ \vspace{0.5cm}
\hfill4th MaxPool\hfill~~~~~\hfill4th MaxPool\hfill~~~~~\hfill4th MaxPool\hfill~\\ \vspace{0.3cm}
\includegraphics[width=3.9cm]{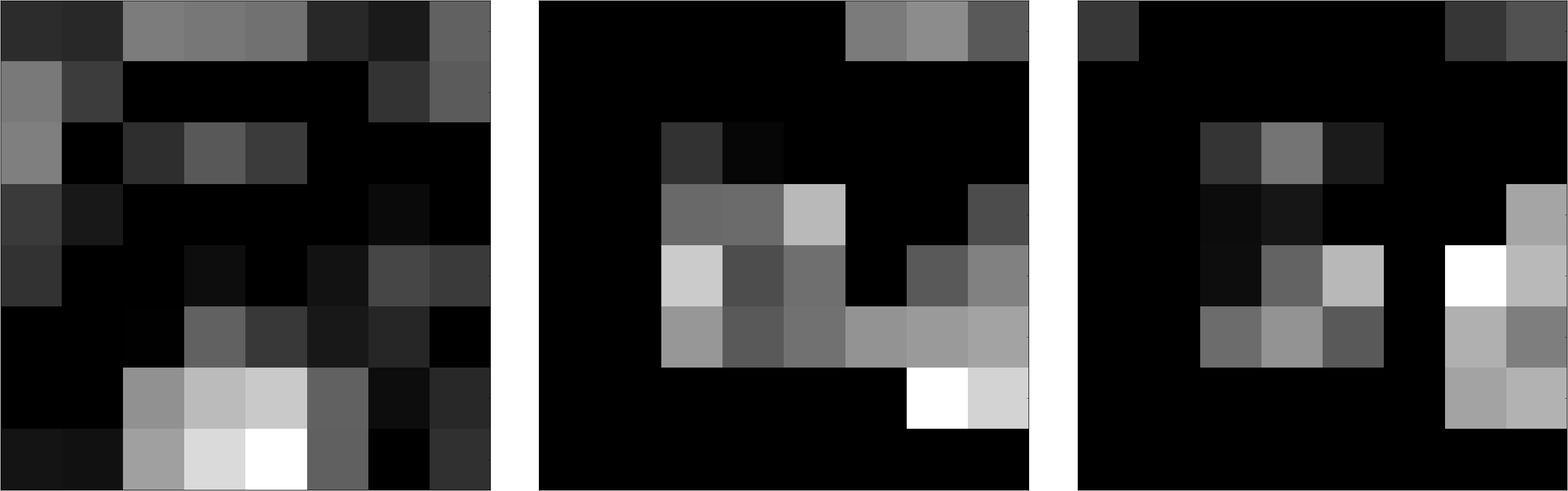}\hfill\includegraphics[width=3.9cm]{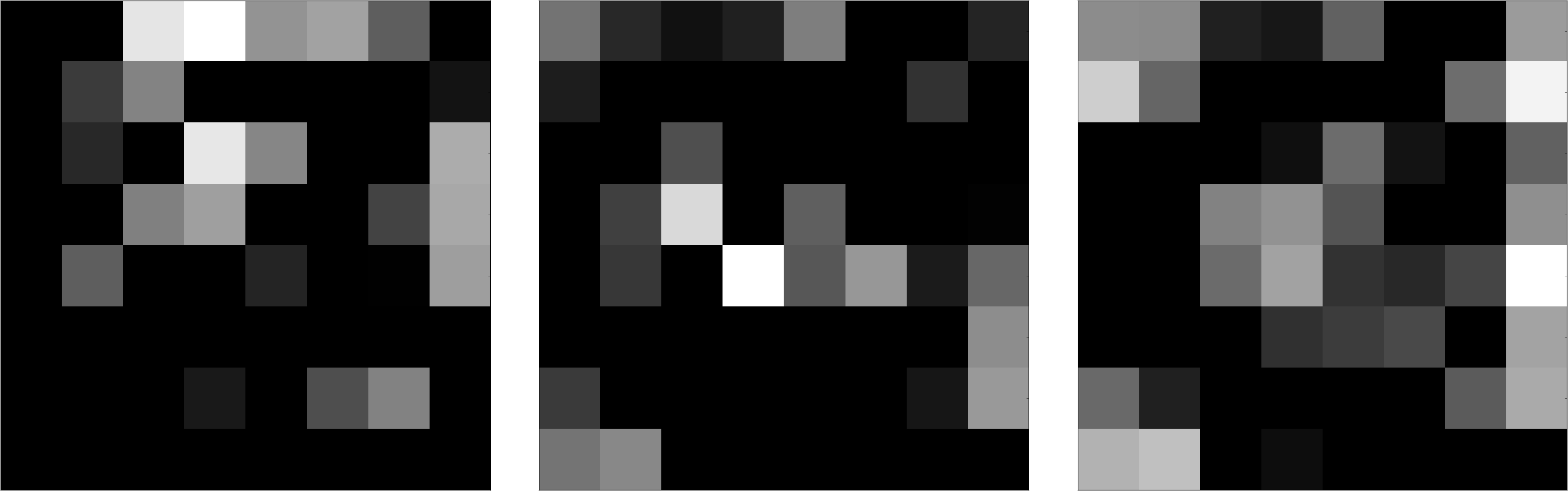}\hfill\includegraphics[width=3.9cm]{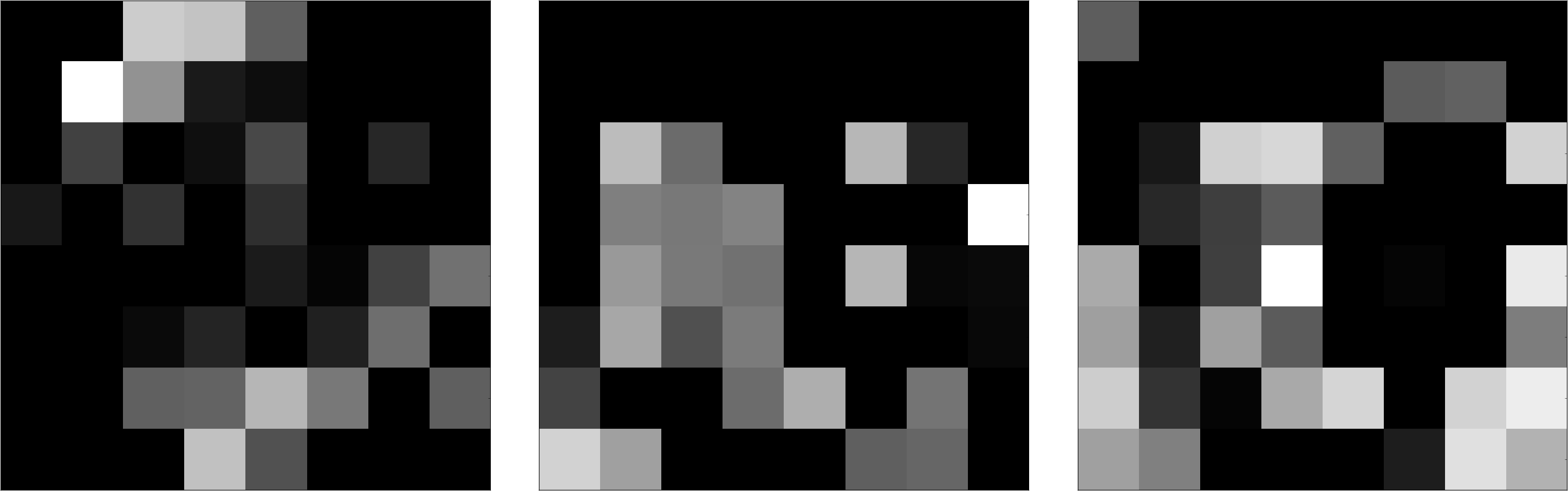}\\ \vspace{0.1cm}
\hfill(a)\hfill~~~~~\hfill(b)\hfill~~~~~\hfill(c)\hfill~\\ \vspace{0.2cm}

\caption{Visualization of the activations for three selected filters in each max pooling layer of our CNN for three different test images, (a)-(b) the first two from a same person and (c) the third one from a different subject. In our interpretation, the first column of activations in all of them illustrate the perception of presence/absence of earrings. As may be observed, even for images from the same subject (a) and (b), the first column has more intense activations in the bottom part of the image when there is an earring in the image. The second and third columns for each image were used to show different concepts learned in different depths of the network. The first max pooling layer usually contains low level features, such as the vertical and horizontal edges. In the second layer, we start encountering more complex concepts, such as helix contours and borders from specific ear parts. As we go deeper in network, it becomes harder to interpret the meaning of the features, although we can always find some activations concentrated in specific parts of the ear, such as concha contours in the third layer and internal ear parts in the fourth layer.}
\label{fig_cnnvis}
\end{figure}

This CNN outputs 512-dimensional descriptors that can be matched through the cosine distance, making the entire processing time ({\it i.e.} description and matching) comparable to that of handcrafted descriptors presented in Section~\ref{sec_handcrafted}. For a given training set, the network optimization was performed in batches of 128 images for 1000 epochs using softmax loss only, and then the weighted sum of softmax and center losses was used until convergence was reached ({\it i.e.} no improvement after 50 epochs). As in Section~\ref{sec_landmark}, we performed data augmentation operations to increase the number of training images: random rotation between $-10^\circ$ and $10^\circ$; random crop with 85\% to 100\% of the original image size; random contrast change increasing or decreasing the range of  pixel intensities in up to 50\%. To help understand what kind of patterns are being encoded by the CNN, we provide a visual analysis of some of the learned filters in Figure~\ref{fig_cnnvis}.

\section{Score fusion}
\label{sec_fusion}

There are different kinds of multimodal systems that address problems associated with single modality systems~\cite{Jain2004}, but a multimodal system based on multiple matchers is the most adequate one for wild scenarios. The reason is that it is not always possible to have multiple biometric traits ({\it e.g.} face and ear), multiple units of a biometric trait ({\it e.g.} thumb and index fingerprints) or multiple samples of the same biometric trait ({\it e.g.} face in video), but we can always apply multiple matching techniques to a single biometric sample.

In order to fuse matchers based on the descriptors previously presented, we evaluated different fusion schemes at score level, such as sum, min, max and product rules, and ended up using the sum rule~\cite{Kittler1998} as it achieved the best results in our experiments. Before fusion, score normalization is carried out considering an identification scenario, where the only scores available at a single time are the ones between the probe and all gallery images. To this end, we discover the minimum and maximum score values and then perform a min-max normalization~\cite{Jain2005}.

\section{Experimental results}

We designed our experiments to validate each module of our recognition framework. Thus, in the following sections we present separate results for landmark detection, geometric normalization, CNN-based description and descriptor fusion. We also compare our results to the state-of-the-art when possible.

\subsection{Landmark detection results}

\begin{figure}[!ht]
\centering
\includegraphics[width=12.0cm]{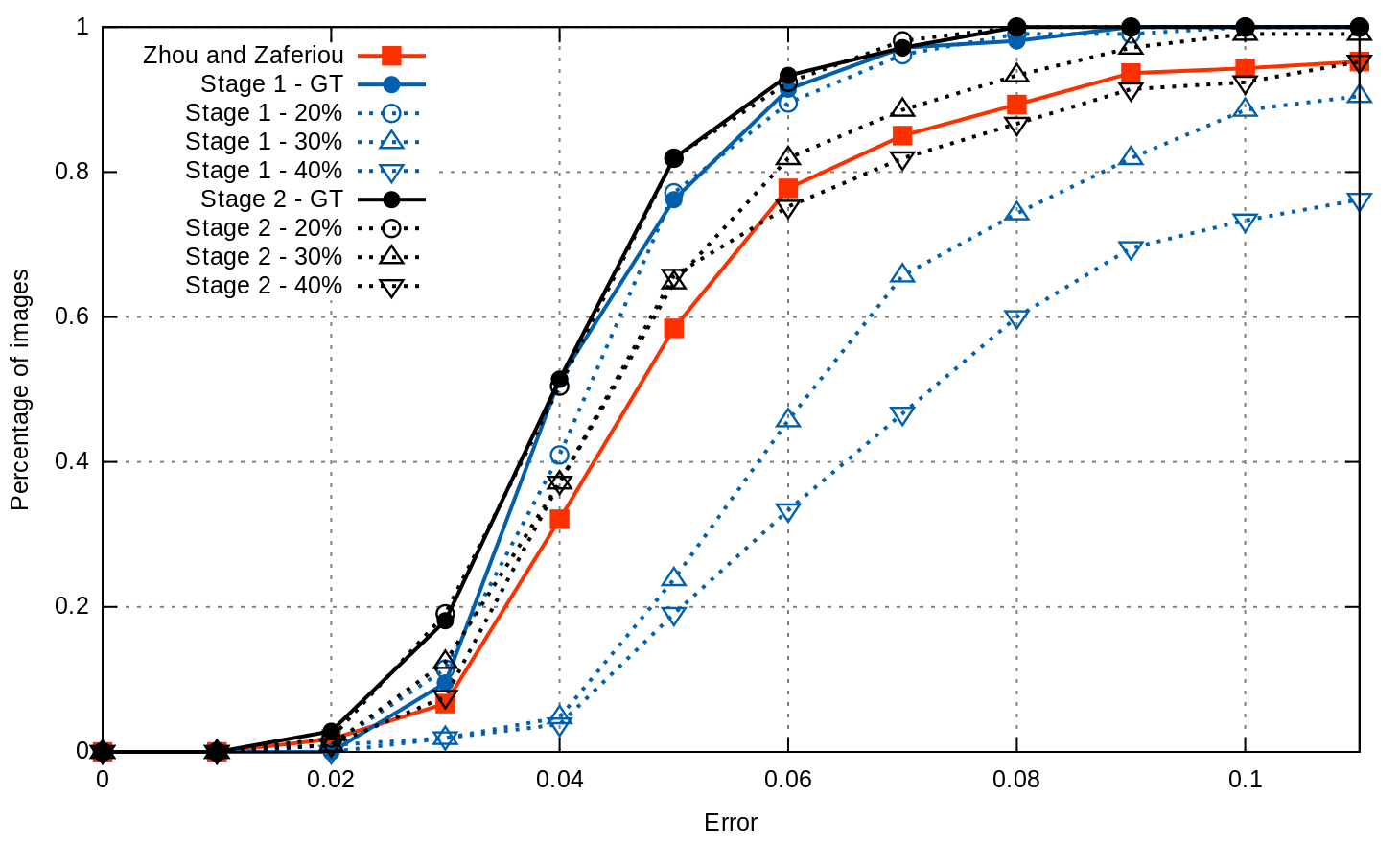}
\caption{Cumulative error distribution for landmark detection using the proposed approach and Zhou~and~Zaferiou's approach~\cite{Zhou2017}.}
\label{fig_landmark}
\end{figure}

Zhou~and~Zaferiou~\cite{Zhou2017} evaluated different variations of Active Appearance Models (AAM) using the test set from ITWE's Collection A. Their best result was achieved by training a holistic AAM based on SIFT features. As initialization for their landmark detector, they used a HOG-based ear detector. They computed the cumulative error distribution using the test set of the same database, where the error for an image is the normalized point-to-point error with respect to the diagonal of the bounding box for the ground truth annotations. Their best result is shown as a line with solid squares in Figure~\ref{fig_landmark}.

We performed the same evaluation for the proposed landmark detector in four different scenarios. In the first one, the ground truth annotations were used to obtain the center and the size of the ear. This reflects the performance of our method in a perfect scenario, in which ear's location and size are reliably retrieved by an ear detector. Then, to simulate scenarios in which the ear detector does not perform that well, we added random variations with up to 20\%, 30\% and 40\% of the ear size to the ground truth values of the first scenario. Results for these four scenarios are also shown in Figure~\ref{fig_landmark}. As may be observed, the two-stage landmark detector performs slightly better than the single-stage one when using up to 20\% of variation, and there is no significant difference in performance between ground truth initialization and an initialization with 20\% of variation. This is expected, as this amount of variation was taken into account during the training stage. For larger variations that are unknown to the training, a single-stage landmark detector can have a considerable drop in performance, but our two-stage solution does not experience a considerable drop.  It is able to perform at least as well as the state-of-the-art~\cite{Zhou2017}.

\subsection{Normalization results}

Since there is no normalization ground truth, we evaluated the benefits of the normalization process described in Section~\ref{sec_normalization} by checking the difference in the recognition performance with and without normalization for different handcrafted descriptors. To this end, we normalized all images from the AWE database and followed the same protocol proposed by Emersic~\emph{et~al.}~\cite{Emersic2017} that was released in their toolbox.  We used the development set that contains 60\% of the AWE images. We performed a 10-fold cross-validation, and report the mean accuracy and standard deviation in Table~\ref{table-handcrafted}. It is worth saying that even images that were not correctly normalized were still used in all recognition experiments, and that all folds are exactly the same in our and Emersic~\emph{et~al.}'s works. Table~\ref{table-handcrafted} shows Emersic~\emph{et~al.}'s results, our reproduction of their experiments and our results using normalized images in terms of Rank-1 and Equal Error Rate(EER). It is possible to observe that our results without normalization are very similar to the ones reported by Emersic~\emph{et~al.}, showing that our reproduction of their experiments was successful, and that our results with normalization obtained higher Rank-1 recognition rates for all features and a lower EER in most of the cases. These results illustrate that an effective normalization approach can help improve performance when using description techniques that are not necessarily robust to wild ear variations.

\begin{table}[!h]
\caption{Rank-1 and EER results for the AWE database as reported by Emersic~\emph{et~al.}~\cite{Emersic2017} and as in our reproduction of their experiments using images with (norm) and without (raw) normalization.\label{table-handcrafted}}{
\footnotesize
\begin{tabular}{|c|c|c|c|c|c|c|} \hline
 & \multicolumn{2}{c|}{\textbf{Emersic~\emph{et~al.}'s work}} & \multicolumn{2}{c|}{\textbf{This work (raw)}} & \multicolumn{2}{c|}{\textbf{This work (norm)}} \\ \hline
\textbf{Method} & \textbf{Rank-1} & \textbf{EER} & \textbf{Rank-1} & \textbf{EER} & \textbf{Rank-1} & \textbf{EER} \\ \hline
LBP   & 43.5$\pm$7.1 & 32.0$\pm$7.4 & 43.5$\pm$7.1 & 32.3$\pm$2.2 & \bf 50.5$\pm$6.8  & \bf 29.8$\pm$2.2 \\
BSIF  & 48.4$\pm$6.8 & 30.0$\pm$9.6 & 48.4$\pm$6.4 & \bf 30.2$\pm$2.9 & \bf 53.1$\pm$7.8  & 30.8$\pm$2.7 \\
LPQ   & 42.8$\pm$7.1 & 30.0$\pm$7.4 & 42.6$\pm$7.0 & 31.7$\pm$2.7 & \bf 47.8$\pm$8.9  & \bf 29.7$\pm$3.4 \\
RILPQ & 43.3$\pm$9.4 & 34.0$\pm$6.4 & 43.5$\pm$9.2 & 34.3$\pm$3.2 & \bf 46.4$\pm$8.4  & \bf 33.8$\pm$4.3 \\
POEM  & 49.6$\pm$6.8 & 29.1$\pm$9.1 & 49.6$\pm$6.8 & \bf 29.8$\pm$2.6 & \bf 54.3$\pm$7.8  & 30.6$\pm$4.3 \\
HOG   & 43.9$\pm$7.9 & 31.9$\pm$7.8 & 48.1$\pm$8.8 & 30.5$\pm$2.1 & \bf 57.1$\pm$8.1  & \bf 26.8$\pm$3.6 \\
DSIFT & 43.4$\pm$8.6 & 30.9$\pm$8.4 & 42.2$\pm$9.0 & \bf 33.1$\pm$2.8 & \bf 45.8$\pm$10.3 & 33.2$\pm$3.1 \\
Gabor & 39.8$\pm$7.1 & 32.1$\pm$8.1 & 39.8$\pm$7.1 & 31.7$\pm$3.8 & \bf 42.6$\pm$6.2  & \bf 30.7$\pm$3.4 \\ \hline
\end{tabular}
}{}
\end{table}

\subsection{CNN description results}
\label{sec-rescnn}

In order to learn features for the problem of recognizing ears in the wild, we divided each of the IIT, WPUTE, AWE and ITWE databases in two sets, one for training and one for testing. The division was conducted in a subject-independent way ({\it i.e.} no subject has images in both training and testing sets) by taking the first half of the subjects rounded up and using their images for training, and using the remaining ones for testing. After the automatic normalization process, images in databases with both left and right ears were flipped in a way that all ears had the same orientation. Finally, each image in the training set was transformed into a set of 20 modified images during the data augmentation stage, and we trained five different descriptors: one for the training set of each chosen database, and one using all these training sets combined. We evaluated the EER performance of these five descriptors by performing an all-versus-all comparison in all testing sets available and the results are presented in Table~\ref{table-crossmatching}. As may be observed, the best performance for all unconstrained testing sets was obtained by the descriptor learned using all training sets, followed by the descriptor learned using the training set from the same database. This shows that every database has different types of variations that tend to be overrepresented in models learned from a single database. When all databases are combined, the model benefits from both a wider training set ({\it i.e} more subjects) and less database overfitting. The models do not appear to be overfitting the unconstrained images, as the performance for the IIT test set is about 2\% for all models.

\begin{table}[!h]
\caption{EER results for all testing sets using descriptors learned from each database or from all databases combined. Each row represents a different CNN descriptor and each column shows the accuracy for a specific database.\label{table-crossmatching}}{
\begin{tabular}{|c|c|c|c|c|c|c|} \hline
 & \multicolumn{6}{c|}{\textbf{TEST}} \\ \hline
\textbf{TRAIN} & IIT        & WPUTE       & WPUTE$^*$  & AWE         & AWE$^*$     & ITWE        \\ \hline
IIT            & \bf 1.76\% & 29.62\%     & 25.85\%    & 35.29\%     & 33.56\%     & 35.47\%     \\
WPUTE          & 2.12\%     & 15.95\%     & \bf 9.40\% & 29.87\%     & 28.33\%     & 29.46\%     \\
AWE            & 2.12\%     & 25.03\%     & 20.04\%    & 26.53\%     & 23.52\%     & 25.68\%     \\
ITWE           & 2.37\%     & 23.50\%     & 18.93\%    & 27.51\%     & 25.30\%     & 22.09\%     \\
ALL            & 2.59\%     & \bf 15.17\% & 9.59\%     & \bf 25.42\% & \bf 22.93\% & \bf 19.68\% \\ \hline
\end{tabular}
}{\\$^*$ As WPUTE and AWE distinguish left and right ears, we also show results considering only genuine matchings between ear images from the same side of the head.}
\end{table}

We also show in Table~\ref{table-crossmatching} that knowing whether or not the image is of the left ear or right ear is helpful during the recognition process. If we only consider genuine matchings as the matchings between ear images from the same side of the head, the EER is reduced in about 4-6\% for the WPUTE database and in approximately 2-3\% for the AWE database in all tests. These results corroborate the findings of Yan~and~Bowyer~\cite{Yan2005} regarding ear asymmetry, but in an uncontrolled scenario. However, it is not always possible to have this information, so we did not consider ear asymmetry in the following experiments and classified matchings between different ears of the same person as genuine.

Zhou~and~Zaferiou~\cite{Zhou2017} used transfer learning in order to employ CNN descriptors previously trained in a different domain~\cite{Simonyan2014} for ear recognition. To this end, they evaluated both Support Vector Machines (SVM) and Linear Discriminant Analysis (LDA) for matching those descriptors, and achieved about 30\% EER for the ITWE database. Their testing/training proportion was 80\%/20\%, and the division was not made in a subject-independent manner. Even though we considered a more difficult scenario, with a 50\%/50\% testing/training subject-independent split, we still achieve a considerably lower EER in all cases where a true unconstrained database was used for training ({\it i.e.} AWE, ITWE and ALL), as may be observed in Table~\ref{table-crossmatching}. These results show that even when a small number of images are available for the ear domain, it may be worth it to train a domain-specific CNN.

\subsection{Fusion results}

The first round of fusion experiments was performed using the testing set of the AWE database, since this was the most challenging one in our previous experiment. We evaluate the fusion of all possible pairs of features, including all holistic, handcrafted and learned features presented in Section~\ref{sec_recognition}. The chosen CNN model was the one with the best result in Table~\ref{table-crossmatching} (ALL). Table~\ref{table-awe} shows individual results for each feature, as well as the top fusion results in terms of Rank-1, Rank-5, Area Under Curve (AUC) and EER. As may be observed, although learned features and top handcrafted features perform equally well individually for Rank-1 and Rank-5, fusion results are dominated by CNN combinations. We believe this is caused by a larger correlation among handcrafted features, which usually have a similar design inspiration that was exploited in slightly different ways by different experts ({\it e.g} quantize gradients, encode neighbors). Thus, CNN is probably learning something complementary to the experts' knowledge, which is corroborated by the fact that nearly all combinations between CNN and one handcrafted feature perform better than all combinations between two handcrafted features.

\begin{table}[!h]
\caption{Individual and fusion results for all descriptors in Section~\ref{sec_recognition} using the AWE database. Individual results were grouped by descriptor type, and handcrafted features were grouped in to categories, the first one (Handcrafted I) for methods based on neighborhood encoding and the second one (Handcrafted II) for methods based on gradient orientations.\label{table-awe}}{
\begin{tabular}{|c|c|c|c|c|c|} \hline
\textbf{Type} & \textbf{Descriptors} & \textbf{Rank-1} & \textbf{Rank-5} & \textbf{AUC} & \textbf{EER} \\ \hline
Holistic       & PCA              & 43.0\%     & 64.4\%     & 0.866     & 37.87\%     \\ \hline
Handcrafted I  & POEM             & 65.2\%     & 85.0\%     & 0.948     & 31.68\%     \\
               & BSIF             & 63.8\%     & 83.4\%     & 0.939     & 30.82\%     \\
               & LBP              & 62.0\%     & 82.6\%     & 0.939     & 30.00\%     \\
               & LPQ              & 59.6\%     & 84.2\%     & 0.942     & 30.52\%     \\
               & RILPQ            & 55.0\%     & 79.2\%     & 0.926     & 34.07\%     \\ \hline
Handcrafted II & HOG              & 64.2\%     & 86.2\%     & 0.955     & 29.33\%     \\
               & DSIFT            & 57.8\%     & 78.4\%     & 0.916     & 32.99\%     \\
               & GABOR            & 50.2\%     & 75.6\%     & 0.911     & 32.56\%     \\ \hline
Learned        & CNN              & 64.2\%     & 86.2\%     & 0.957     & \bf 22.89\% \\ \hline
Sum fusion     & CNN+HOG          & \bf 75.6\% & \bf 90.6\% & \bf 0.972 & \bf 22.87\% \\
               & CNN+POEM         & 75.4\%     & 90.4\%     & 0.968     & 24.29\%     \\
               & CNN+LPQ          & 72.8\%     & 88.6\%     & 0.966     & 23.61\%     \\
               & CNN+RILPQ        & 72.0\%     & \bf 90.6\% & 0.962     & 25.11\%     \\
               & HOG+BSIF         & 70.8\%     & 88.6\%     & 0.963     & 28.34\%     \\
               & CNN+BSIF         & 70.2\%     & 89.8\%     & 0.963     & 24.18\%     \\
               & CNN+LBP          & 70.0\%     & 89.4\%     & 0.964     & 23.53\%     \\
               & HOG+RILPQ        & 70.0\%     & 86.4\%     & 0.957     & 29.89\%     \\
               & CNN+GABOR        & 69.4\%     & 88.6\%     & 0.963     & 24.56\%     \\
               & HOG+LPQ          & 69.0\%     & 86.4\%     & 0.960     & 28.67\%     \\ \hline
\end{tabular}
}{}
\end{table}

In our second round of fusion experiments, we reproduced two experiments proposed by Emersic~\emph{et~al.}~\cite{Emersic2017b} to evaluate challenge participants through the UERC database, one to evaluate the overall performance and other to evaluate the scalability of the recognition approaches. To this end, we normalize the UERC training images and use them to learn a sixth CNN descriptor ({\it i.e.} data augmentation was used to balance the classes in a way that each subject ended up with 200 images). As UERC test images do not have the same orientation and ground truth annotations are not provided, we also trained a simple side classifier by changing the output size of the network presented in Table~\ref{cnn-landmarks} to two classes (left and right) and then training it for the UERC training images using softmax loss and the Adam optimization algorithm.  Because images of this database are already cropped, the entire testing process was fully automatic.

\begin{figure}[!ht]
\centering
\subfloat[]{\label{fig_fusion1}\includegraphics[width=10.3cm]{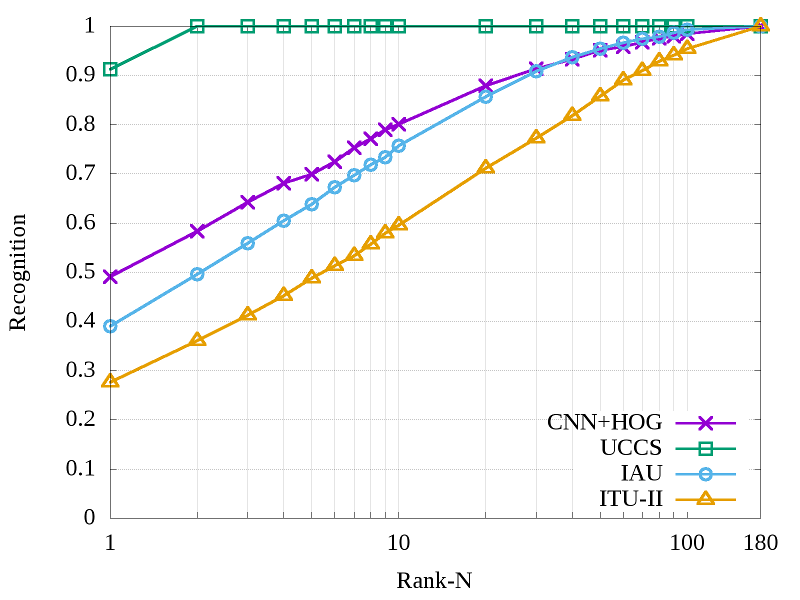}}\hfill
\subfloat[]{\label{fig_fusion2}\includegraphics[width=10.3cm]{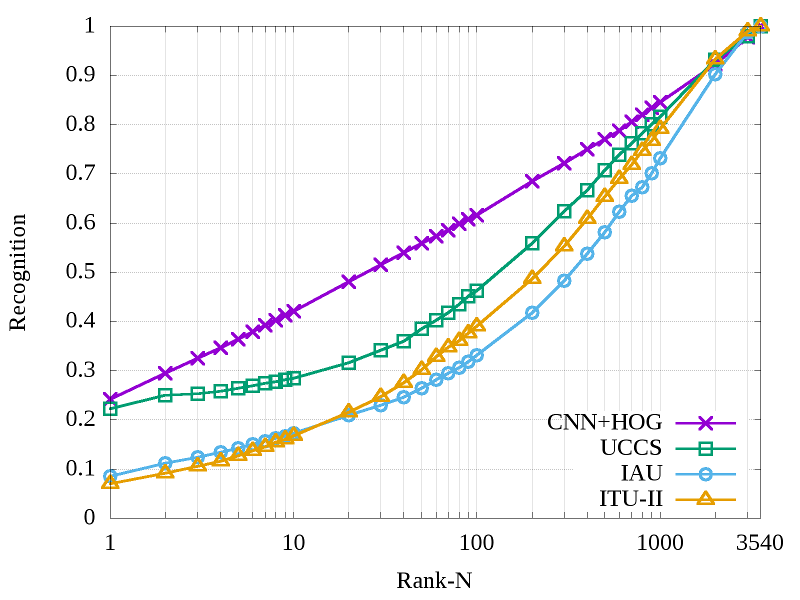}}
\caption{CMC curves for all participants of the UERC challenge plus our best fusion results obtained by combining CNN and HOG considering the (a) overall performance evaluation and (b) scalability evaluation protocols.}
\label{fig_fusion}
\end{figure}

For the overall performance evaluation, only the first 1,800 test images from 180 subjects are used in an all-versus-all comparison. In this experiment, we only use CNN and three other handcrafted features: HOG, POEM and LBP. HOG and POEM obtained the best fusion results with CNN in Table~\ref{table-awe}, and LBP was a baseline approach for participants of the UERC challenge~\cite{Emersic2017b}. Table~\ref{table-uerc} shows individual results for each feature, the fusion results using weighted sum, and the best results reported in the UERC challenge. As may be observed, our normalization resulted in a considerable boost in the performance for handcrafted descriptors.  We achieve more than 20\% improvement in Rank-1 when comparing our LBP result to its baseline version without normalization. Again, individual ranking performances of learned and handcrafted features were similar, but the CNN fusion pairings stood out. Our performance was higher than all participants of the challenge except University of Colorado Colorado Springs (UCCS), whose results we have not verified, as they appear to be using test images for training. CMC curves for the best performing works are presented in Figure~\ref{fig_fusion1}.

\begin{table}[!h]
\caption{Individual and fusion results for CNN, HOG, POEM and LBP in the overall performance evaluation through the UERC protocol, as well as the top scoring participants of the UERC challenge.\label{table-uerc}}{
\begin{tabular}{|c|c|c|c|c|c|} \hline
\textbf{Type} & \textbf{Descriptors} & \textbf{Rank-1} & \textbf{Rank-5} & \textbf{AUC} & \textbf{EER} \\ \hline
Handcrafted I  & POEM                             & 36.83\%     & 58.44\%     & 0.907     & 36.17\%     \\
               & LBP                              & 35.00\%     & 55.11\%     & 0.897     & 35.81\%     \\ \hline
Handcrafted II & HOG                              & 39.78\%     & 60.56\%     & 0.916     & 35.51\%     \\ \hline
Learned        & CNN                              & 36.94\%     & 60.56\%     & 0.930     & \bf 26.77\% \\ \hline
Sum fusion     & CNN+HOG                          & \bf 49.06\% & \bf 69.94\% & \bf 0.951 & 27.84\%     \\
               & CNN+POEM                         & 47.28\%     & \bf 70.00\% & 0.948     & 28.21\%     \\
               & CNN+LBP                          & 45.22\%     & 67.44\%     & 0.946     & 28.05\%     \\
               & HOG+POEM                         & 43.06\%     & 64.33\%     & 0.926     & 35.14\%     \\
               & HOG+LBP                          & 41.22\%     & 60.89\%     & 0.919     & 35.11\%     \\
               & POEM+LBP                         & 38.56\%     & 59.00\%     & 0.911     & 35.39\%     \\ \hline
Literature     & UCCS~\cite{Emersic2017b}         & \it 90.4\%$^*$ & \it 100.0\%$^*$ & \it 0.994$^*$ &  \\
               & IAU~\cite{Emersic2017b}          & 38.5\%      & 63.2\%      & 0.940     &             \\
               & ITU-II~\cite{Emersic2017b}       & 27.3\%      & 48.3\%      & 0.877     &             \\
               & LBP-baseline~\cite{Emersic2017b} & 14.3\%      & 28.6\%      & 0.759     &             \\ \hline
\end{tabular}
}{\\$^*$ These results still require verification.}
\end{table}

For the scalability evaluation, we match all images from subjects with at least two images to all other test images, totaling 7,442$\times$9,499 matching pairs. This experiment increases the number of subjects to 3,540 and also adds many images with poor quality, affecting considerably the performance of the evaluated approaches. In Table~\ref{table-uerc2} we show results for CNN, HOG and POEM, for all possible fusion among two of them, and for the best performing approaches in the UERC challenge. We can see that the combination of CNN and HOG was again the best performing method for lower ranks, and that these results show our approach as the most scalable unconstrained ear recognition approach. CMC curves for the best performing works are presented in Figure~\ref{fig_fusion2} and show how well our approach performs for lower ranks, outperforming all other works by at least 10\% in most ranks before Rank-300.

\begin{table}[!h]
\caption{Individual and fusion results for CNN, HOG and POEM in the scalability evaluation through the UERC protocol, as well as the top scoring participants of the UERC challenge.\label{table-uerc2}}{
\begin{tabular}{|c|c|c|c|c|c|} \hline
\textbf{Type} & \textbf{Descriptors} & \textbf{Rank-1} & \textbf{Rank-5} & \textbf{AUC} & \textbf{EER} \\ \hline
Handcrafted I  & POEM                             & 17.98\%     & 28.48\%     & 0.851     & 40.38\%     \\ \hline
Handcrafted II & HOG                              & 18.50\%     & 28.78\%     & 0.851     & 40.80\%     \\ \hline
Learned        & CNN                              & 17.13\%     & 28.73\%     & 0.873     & 35.92\%     \\ \hline
Sum fusion     & CNN+HOG                          & \bf 24.17\% & \bf 36.43\% & 0.881     & 36.24\%     \\
               & CNN+POEM                         & 23.02\%     & 35.70\%     & \bf 0.882 & \bf 35.82\% \\
               & HOG+POEM                         & 20.57\%     & 32.12\%     & 0.856     & 40.26\%     \\ \hline
Literature     & UCCS~\cite{Emersic2017b}         & \it 22.3\%$^*$ & \it 26.3\%$^*$ & \it 0.858$^*$ &   \\
               & IAU~\cite{Emersic2017b}          & 8.4\%       & 14.2\%      & 0.810     &             \\
               & ITU-II~\cite{Emersic2017b}       & 6.9\%       & 12.8\%      & 0.844     &             \\ \hline
\end{tabular}
}{\\$^*$ These results still require verification.}
\end{table}

Table~\ref{table-uerc2} also shows that our CNN outperformed the other two top-scoring CNN-based approaches proposed by researchers from the Islamic Azad University (IAU) and the Istanbul Technical University (ITU-II)~\cite{Emersic2017b}. Similarly to Zhou~and~Zaferiou~\cite{Zhou2017}, IAU and ITU-II employed transfer learning approaches in a network from a different domain~\cite{Simonyan2014} and were not able achieve results as high as our domain-specific CNN.

\subsection{Discussion}

Unconstrained ear recognition is a very challenging problem, and recent efforts to provide data for unconstrained ear images for research are helpful. Initial databases such as IIT and WPUTE were captured images instead of wild images. They do not have much intraclass and interclass variations. The initial wild databases collected from the Internet such as AWE and ITWE still lack interclass variability due to their small number of subjects. The UERC database is a vast repository with thousands of subjects and images with intraclass variations and interclass variations.  It is the most challenging ear dataset that we are aware of.

Although initial ear recognition works have consistently used ear alignment before recognition~\cite{Chang2003,Kumar2012}, researches for unconstrained ear recognition were mostly focused in finding robust features~\cite{Emersic2017}. Even among the UERC participants, only the Imperial College London (ICL) has used an alignment step, although they used an AAM-based solution~\cite{Zhou2017} that may not be as successful for wild images as recent techniques such as CNNs (see Figure~\ref{fig_landmark}). Nevertheless, we attribute a big part of the success in our results to the normalization step. It considerably increased the performance of traditional methods, such as handcrafted features in Tables~\ref{table-handcrafted}~and~\ref{table-uerc}, and also helped the deep learning process by letting it focus on what matters the most for the recognition task. It also helped in our cross-dataset experiments shown in Table~\ref{table-crossmatching}, as we do not have problems with different cropping areas or noise in ear location.

Our CNN descriptors were comparable to the best handcrafted descriptors in terms of Rank-N results, but they performed better in terms of EER in all experiments, meaning that they were more accurate for verification purposes. In addition, our performance was favorably compared to the best performing participants of the UERC challenge, as shown in Table~\ref{table-uerc2}, and to other state-of-the-art work in Section~\ref{sec-rescnn}. There were two factors that may have helped to achieve these results: we trained CNNs from scratch specifically to our problem domain, and we used a discriminative learning technique based on center loss that was proposed by Wen~\emph{et~al.}~\cite{Wen2016}.

Finally, as learned and handcrafted features were achieving similar ranking results for our normalized images, we decided to combine them through score fusion in order to pursue a better performance. We ended up discovering that the combination of our CNN descriptors and handcrafted descriptors achieved much better results in all experiments. None of the combinations between a pair of handcrafted features could get close to the top scores, which may be explained by the fact that handcrafted features are highly correlated due to their similar design. On the other hand, CNN descriptors do not follow an expert's design and are probably learning some discriminative information that is complementary to most handcrafted descriptors, as may be observed in Tables~\ref{table-awe},~\ref{table-uerc}~and~\ref{table-uerc2}.

\section{Conclusion}

Unconstrained ear recognition is a very challenging problem. To address this challenge, we provide a framework that combines handcrafted features and CNN. We test our framework using the most challenging publicly available ear database that we are aware of.  Our results are considerably better than recently published works and are less impacted by database scale. We gain invaluable lessons that can further enhance unconstrained ear recognition research:

\begin{itemize}
\item handcrafted features are not dead; this is consistent with the finding in action recognition works~\cite{Zhu2016};\\
\item handcrafted features and CNN are complementary;\\
\item normalization is critical and enhances performance recognition of handcrafted features;\\
\item CNN combined with any of the aforementioned descriptors improves recognition.
\end{itemize}

There is still a lot to be learned in order to address the issues with unconstrained ear recognition.  Our work demonstrates that CNN and handcrafted features are a good starting point. 

\bibliographystyle{unsrt}
\bibliography{example}

\end{document}